\documentclass[]{interact}

\usepackage{epstopdf}% To incorporate .eps illustrations using PDFLaTeX, etc.
\usepackage{subfigure}% Support for small, `sub' figures and tables
\usepackage{multirow}

\usepackage{natbib}% Citation support using natbib.sty
\bibpunct[, ]{(}{)}{;}{a}{}{,}% Citation support using natbib.sty
\renewcommand\bibfont{\fontsize{10}{12}\selectfont}% Bibliography support using natbib.sty

\theoremstyle{plain}% Theorem-like structures provided by amsthm.sty

\theoremstyle{definition}

\theoremstyle{remark}

\usepackage{booktabs} % For formal tables
\usepackage{url}
\usepackage{amsmath}
\usepackage{float}
\usepackage{subfig}
\usepackage{graphicx}
\usepackage{color}
\usepackage[noend]{algorithmic}
\usepackage{algorithm}
\usepackage{multirow}
\usepackage{verbatim}
\usepackage{amsmath}
\usepackage{amsthm}
\usepackage{indentfirst}
\usepackage{breqn}
\usepackage{multicol,lipsum}
\usepackage{amssymb} % \leadsto
\usepackage{cuted}
\usepackage{makecell}

\DeclareMathOperator*{\argmax}{arg\,max}

\renewcommand{\thesubfigure}{(\textit{\alph{subfigure}})}

\begin{document}

%\articletype{ARTICLE TEMPLATE}

\title{Flood Extent Mapping based on High Resolution Aerial Imagery and DEM: A Hidden Markov Tree Approach}

\author{
\name{Zhe Jiang\thanks{CONTACT Zhe Jiang. Email: zjiang@cs.ua.edu} and Arpan Man Sainju}
\affil{Department of Computer Science, the University of Alabama, Tuscaloosa, Alabama, USA}
}

\maketitle

\begin{abstract}

Flood extent mapping plays a crucial role in disaster management and national water forecasting. In recent years, high-resolution optical imagery becomes increasingly available with the deployment of numerous small satellites and drones. However, analyzing such imagery data to extract flood extent poses unique challenges due to the rich noise and shadows, obstacles (e.g., tree canopies, clouds), and spectral confusion between pixel classes (flood, dry) due to spatial heterogeneity. Existing machine learning techniques often focus on spectral and spatial features from raster images without fully incorporating the geographic terrain within classification models. In contrast, we recently proposed a novel machine learning model called geographical hidden Markov tree that integrates spectral features of pixels and topographic constraint from Digital Elevation Model (DEM) data (i.e., water flow directions) in a holistic manner. This paper evaluates the model through case studies on high-resolution aerial imagery from National Oceanic and Atmospheric Administration (NOAA) National Geodetic Survey together with DEM. Three scenes are selected in heavily vegetated floodplains near the cities of Grimesland and Kinston in North Carolina during Hurricane Matthew floods in 2016. Results show that the proposed hidden Markov tree model outperforms several state of the art machine learning algorithms (e.g., random forests, gradient boosted model) by an improvement of F-score (the harmonic mean of the user's accuracy and producer's accuracy) from around 70 to 80\% to over 95\% on our datasets. \end{abstract}

\begin{keywords}
Remote sensing; flood mapping; topography; tree canopies; spatial structured prediction; hidden Markov tree
\end{keywords}

\section{Introduction}\label{sec:intro}

Flood extent mapping plays a crucial role in addressing grand societal challenges such as disaster management,  national water forecasting, as well as energy and food security (\cite{eftelioglu2016spatial}). For example, during Hurricane Harvey floods in 2017, first responders needed to know where flood water was in order to plan rescue efforts to residents in vulnerable communities, to understand the extent of damage to critical infrastructures (e.g., chemical plants and oil refineries), and evaluate the impact on road networks and transportation. In national water forecasting, \cite{nwm} and \cite{iwrss} introduce a Super-resolution National Water Model being operated at the National Water Center, which can forecast the flow of over 2.7 million rivers and streams through the entire continental U.S.. One main issue of the national water model is that it has a large number of parameters in physical models that can only be calibrated and validated based on observations from a few thousands of river gauges. Detailed flood extent maps derived from remote sensing observations provide an alternative way to calibrate and validate the national water model in broad spatial coverage.

%In current practice, flood extent mapping can be generated by flood forecasting models, whose accuracy may be insufficient in high spatial details. Other ways to generate flood maps involve sending a field crew on the ground to record high-water marks, or visually interpreting earth observation imagery (\cite{brivio2002integration}). However, the process is both expensive and time consuming. With the large amount of high-resolution earth imagery being collected from satellites (e.g., DigitalGlobe, Planet Labs), aerial planes (e.g., NOAA National Geodetic Survey), and unmanned aerial vehicles, the cost of manually labeling flood extent becomes prohibitive.
%optical imagery, radar, lidar topography; index based (NDVI), machine learning approach, recent deep learning approach;

Mapping flood extent from earth observation imagery has been extensively studied in the remote sensing community. Existing techniques can be categorized by the types of remote sensors (optical sensor, radar, Light Detection and Ranging (LiDAR)) or the underlying approaches (machine learning, non-machine learning). \cite{fayne2017flood} uses the Normalized Difference Vegetation Index (NDVI) from Moderate Resolution Imaging Spectroradiometer (MODIS) optical imagery together with change detection to classify flood extent. \cite{mcfeeters1996use} and \cite{xu2006modification} propose the use of Normalized Difference Water Index (NDWI) and modified NDWI to detect surface water features respectively.  \cite{sun2011deriving} proposes to use decision trees to map flood extent from MODIS optical images.  \cite{zhang2014blending} proposes the blend of Land Remote-Sensing Satellite (Landsat) and MODIS optical images for urban flood mapping. 
\cite{lin2019improvement} proposes algorithms filter noises from terrain shadows to improve National Aeronautics and Space Administration (NASA)/MODIS Near Real-Time (NRT) Global Flood Mapping.  
\cite{feng2015urban} uses random forest algorithms on Unmanned Aerial Vehicles (UAV) imagery for mapping flood extent in China. One major limitation of using imagery from optical sensors is that the scene is often obscured by obstacles (e.g., cloud, tree canopies). Because of this reason, radar sensor data are often used to identify flood extent since the signals can penetrate through clouds and some vegetated areas. \cite{giustarini2012change} proposes a change detection method on radar imagery from TerraSAR-X. \cite{horritt2003waterline} proposes to use airbone Synthetic Aperture Radar (SAR) imagery to identify flooded vegetation. \cite{townsend2001mapping} proposes to use multi-temporal SAR imagery to map seasonal flooding in forested wetlands. \cite{oberstadler1997assessment} assessed the capabilities of European Remote Sensing satellite ERS-1 SAR data for flood mapping. 
\cite{yang2014change} proposes change detection algorithms for high-resolution SAR images in the context of flood mapping.
Though radar data have the advantage of penetration through clouds and some vegetation, the signals are often very noisy (e.g., speckles due to interference with irregular surface). There are works that propose the integration of optical or radar sensor data together with topographic information from Digital Elevation Model (DEM) to delineate flood extent. 
\cite{wang2002efficient} proposes the use of Landsat optical imagery together with topography from DEM to map flood extent. The approach showed significant improvements in identifying the underestimated flood extent  in heavily vegetated area, but the approach is not automatic and involves manually thresholding the DEM. Such threshold may vary from one region to another. \cite{matgen2007integration} proposes the integration of SAR data, high-precision topographic features, and  a river flow model for near real-time flood management. 
\cite{rahman2019rapid} proposes to use model-driven soil moisture from the Soil Moisture Active Passive (SMAP) data product for rapid flood progress monitoring to estimate crop loss in several flood events, including the Hurricane Harvey flood in Houston 2017. 
There are other review articles on machine learning or data science techniques for earth observation imagery data, including~\cite{jiang2017spatialbook},~\cite{jiang2018survey},~\cite{shekhar2015spatiotemporal},~\cite{jiang2020spatial}, and~\cite{karpatne2016monitoring}. \cite{rahman2017state} provides a review on existing methods and approaches that overcome the constraints in the application of spaceborne remote sensing in flood management, such as bad weather condition and cloud cover.
% {\color{red}existing remote sensing paper for flood mapping}
% The idea is to use a small set of manually collected ground truth (flood and dry locations) in one earth imagery to learn a classification model. Then the model can be used to classify flood pixels in other imagery where ground truth is not available. 

Mapping flood  from high-resolution imagery in heavily vegetated areas poses several unique challenges that are not well addressed in traditional machine learning classification models. First, high-resolution earth imagery often has noise, shadows, clouds, and tree canopies (\cite{jiang2012learning,jiang2013focal,jiang2015focal}). Second, class confusion exists due to heterogeneous spectral features. For example, the spectral signature of pixels in shadows and flooded areas can be very similar but they are in different classes (\cite{jiang2017spatialensemble,jiangtist19}). As another example, pixels of tree canopies overlaying flood water have the same spectral features with pixels of tree canopies in dry areas. 
This is a challenge for machine learning models which classify pixel types largely based on pixel or neighborhood features on the imagery. 
\cite{shen2019inundation} finds that though the automation and robustness of non-obstructed inundation mapping have been achieved with acceptable accuracy, they are not yet satisfactory for the detection of beneath-vegetation flood mapping. 
Third, implicit directed spatial dependency exists between flood pixel locations. Specifically, due to gravity, flood water tends to flow to nearby lower locations following topography (\cite{sainju2020spatial}). Such dependency is not uniform in all directions (anisotropic). Such topographic constraints are often ignored in the existing machine learning image classification algorithms. Finally, the data volume is huge in high-resolution imagery (e.g., hundreds of millions of pixels), requiring algorithms to be computationally scalable.

To address these challenges, a novel spatial classification model called \emph{geographical hidden Markov tree }(HMT) was recently proposed in the data mining community by~\cite{Xie2018,jiang2019hidden,jiang2019geographical}. Intuitively, HMT is a spatial machine learning model that integrates local likelihood of pixels' classes derived from their own spectral signatures and the topographic constraint between pixels derived from water flow directions. Specifically, HMT is a probablistic graphical model that generalizes the common hidden Markov model (HMM) from a one-dimensional (1D) sequence to a two dimensional (2D) geographical map. The hidden class layer contains nodes (pixels) in a reverse tree structure to represent anisotropic spatial dependency of water flow directions. Each hidden class node has an associated observed feature node for the same pixel (local class likelihood). Such a unique model structure can potentially reduce classification errors due to noise, obstacles, and heterogeneity among spectral features of individual pixels. Efficient algorithms exist for model parameter learning and class inference based on the Expectation-Maximization (EM) algorithm with message propagation on tree traversal. 

The goal of this paper is to introduce this technique to the remote sensing community. The paper first introduces the statistical formulation and intuition of the HMT approach. It provides three detailed case studies on high-resolution optical imagery from National Oceanic and Atmospheric Administration (NOAA) National Geodetic Survey (NGS) collected during Hurricane Matthew floods in North Carolina 2016. Results show that HMT outperforms several state of the art machine learning algorithms in accurately mapping flood extent in heavily vegetated areas. Note that the paper only focuses on machine learning approaches for flood mapping. Other non-machine learning approaches (e.g., thresholding a water index, simple change detection) are beyond the scope. The HMT model can be further tested on datasets collected by different sensors among different areas and in different flood events in future studies. 

\section{Materials and Methods}\label{sec:mm}

\subsection{Dataset Description} 
Our real-world high-resolution aerial imagery datasets were acquired by the NOAA Remote Sensing Division to support homeland security and emergency response requirements after each hurricane disaster. The imagery was downloaded from the NOAA NGS at~\cite{ngs} related to the Hurricane Matthew in 2016. The imagery was originally at 0.5 metre spatial resolution with red, green, and blue bands only. The limited spectral bands in the visible spectrum and the varying illumination conditions in the imagery were the major challenges in utilizing the data for flood mapping. We used this dataset to test the potential of HMT in leveraging topography to address spectral limitations. The imagery for flood disasters only covered a few scattered scenes at river floodplains in North Carolina and South Carolina. The specific spatial coverage of the complete datasets can be found on the Emergency Response Imagery Viewer at~\cite{ngs}. The high-resolution imagery was collected only once during the hurricane flood disaster. 
\begin{figure}
    \centering
    \includegraphics[width=5in]{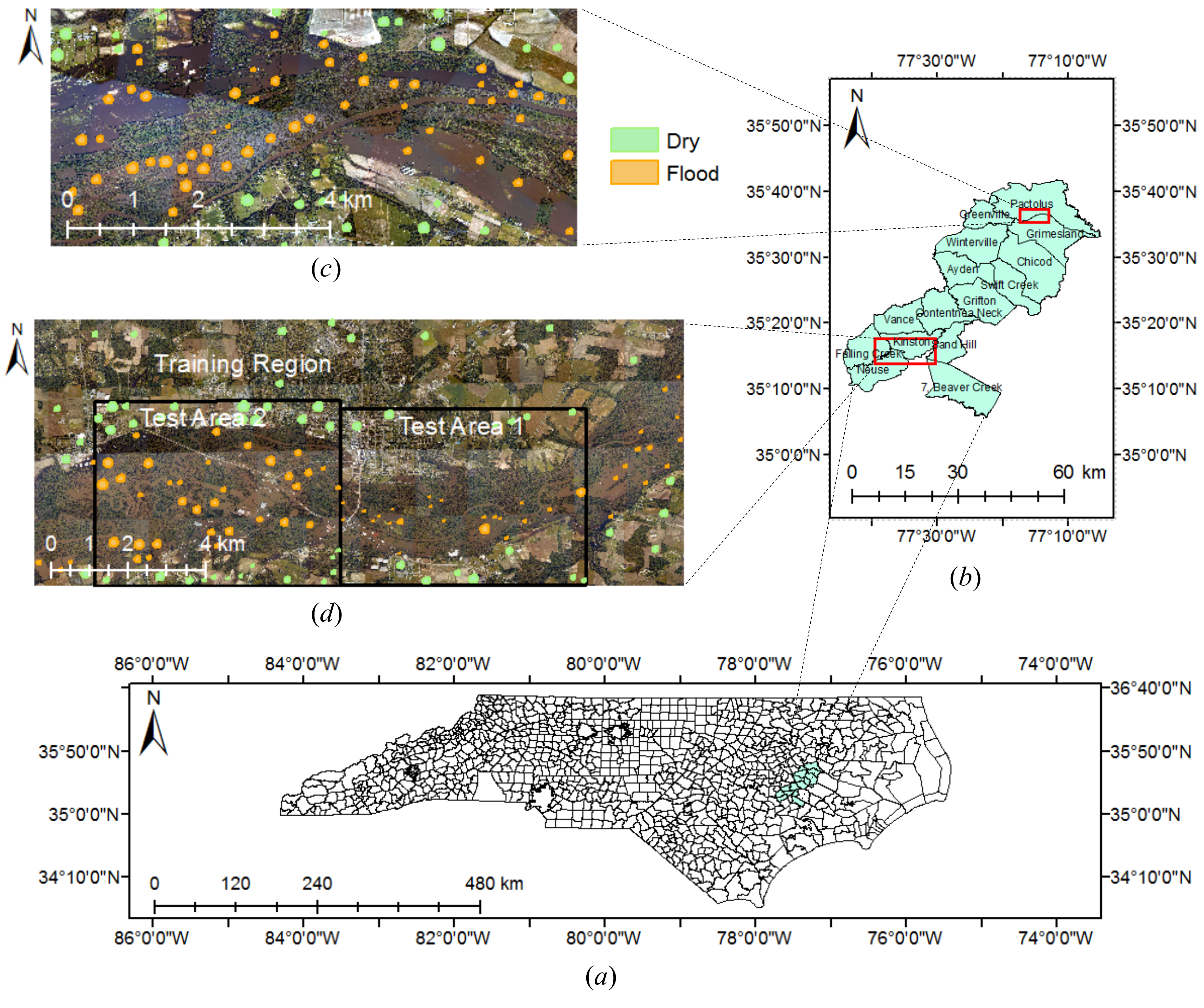}
    \caption{The map of three study areas (Source: NOAA NGS, Date: 15 October 2016). (\textit{a}) North Carolina. (\textit{b}) Study area. (\textit{c}) Grimesland test region. (\textit{d}) Kinston region. The same training set in Kinston region is used for three test scenes (test area 1 and test area 2 in Kinston region as well as Grimesland test region).}
    \label{fig:studyArea}
\end{figure}
Due to the large data volume, we cropped three representative scenes near the cities of Grimesland and Kinston in North Carolina. We also resampled the imagery into 2 metre resolution to reduce the data volume and also to make it consistent with the DEM, which was downloaded from~\cite{ncsudem} at 2 metre resolution. The HMT model uses DEM data to construct a flow dependency tree structure on all pixels in the a region. This tree structure is the core component of the HMT model in parameter learning and class inference, since inferring pixels’ classes are not only based on its own spectral signatures but also the likely classes of other pixels on the topographic surface. The locations of the three scenes are shown in  Figure~\ref{fig:studyArea}. 
% Training samples in Grimesland were drawn outside but nearby the test region. Training samples in the Grimesland dataset were drawn far away from the test regions to test model generalizability. 

The locations of the ground truth polygons in the three study areas are also shown in Figure~\ref{fig:studyArea}. These flood and dry polygons were manually labelled through visual interpreting the aerial imagery together with the DEM. The labeled polygons within the rectangular boundary of the test regions are used for testing. The labeled polygons in Kinston region outside the test areas were used for training. Note that the same training set was used for model learning in all three test regions. For the test region in Grimsland, the training data was far away from it so that we can evaluate the effectiveness of the transductive learning in HMT (initializing model parameters from training data and updating the parameters based on test data that is collected far away). Training and testing samples (pixels) were randomly drawn from the ground truth polygons. Their numbers are listed in Table~\ref{tab:data}.

\begin{table}
\centering
\caption{Dataset description}
\begin{tabular}{lcccc}
\hline
\multirow{2}{*}{Dataset} & \multicolumn{2}{c}{Training set}  & \multicolumn{2}{c}{Testing set} \\ 
&Dry & Flood & Dry & Flood \\ \hline
%  &  & (Dry/Flood) & (Dry/flood) \\
%section1: non-spatial ensemble
Grimesland, NC  & 5,000 & 5,000 & 49,597 & 99,532 \\ \hline
Kinston area 1, NC  & 5,000 & 5,000 & 46,245 & 20,331 \\ \hline
Kinston area 2, NC  & 5,000 & 5,000 & 110,805 & 123,366 \\ \hline
\end{tabular}
\label{tab:data}
\end{table}

\subsection{Methods}

\begin{figure}
    \centering
    \subfigure[]{\includegraphics[width=1.8in]{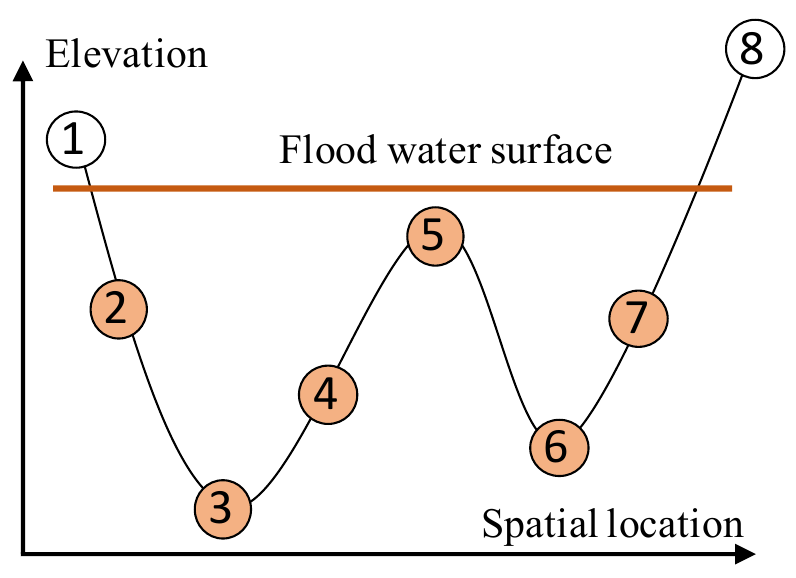}}\hspace{1mm}
    \subfigure[]{\includegraphics[width=1.5in]{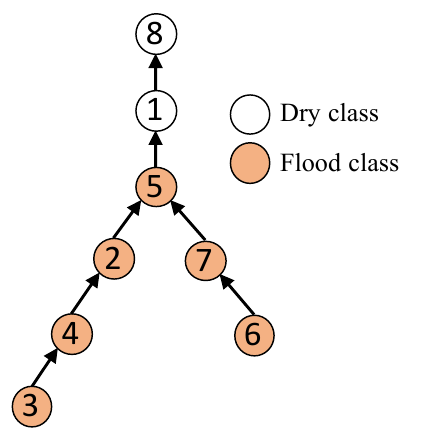}}\hspace{1mm}
    \subfigure[]{\includegraphics[width=1.8in]{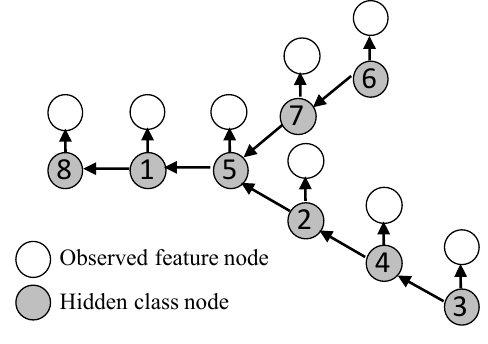}}
    \caption{Illustration of hidden Markov tree framework. (\textit{a}) Eight consecutive sample locations in 1D space. (\textit{b}) Partial order constraint in a reverse tree. (\textit{c}) Hidden Markov tree}
    \label{fig:hmtrt}
\end{figure}

\begin{figure}
    \centering
    \includegraphics[width=5.8in]{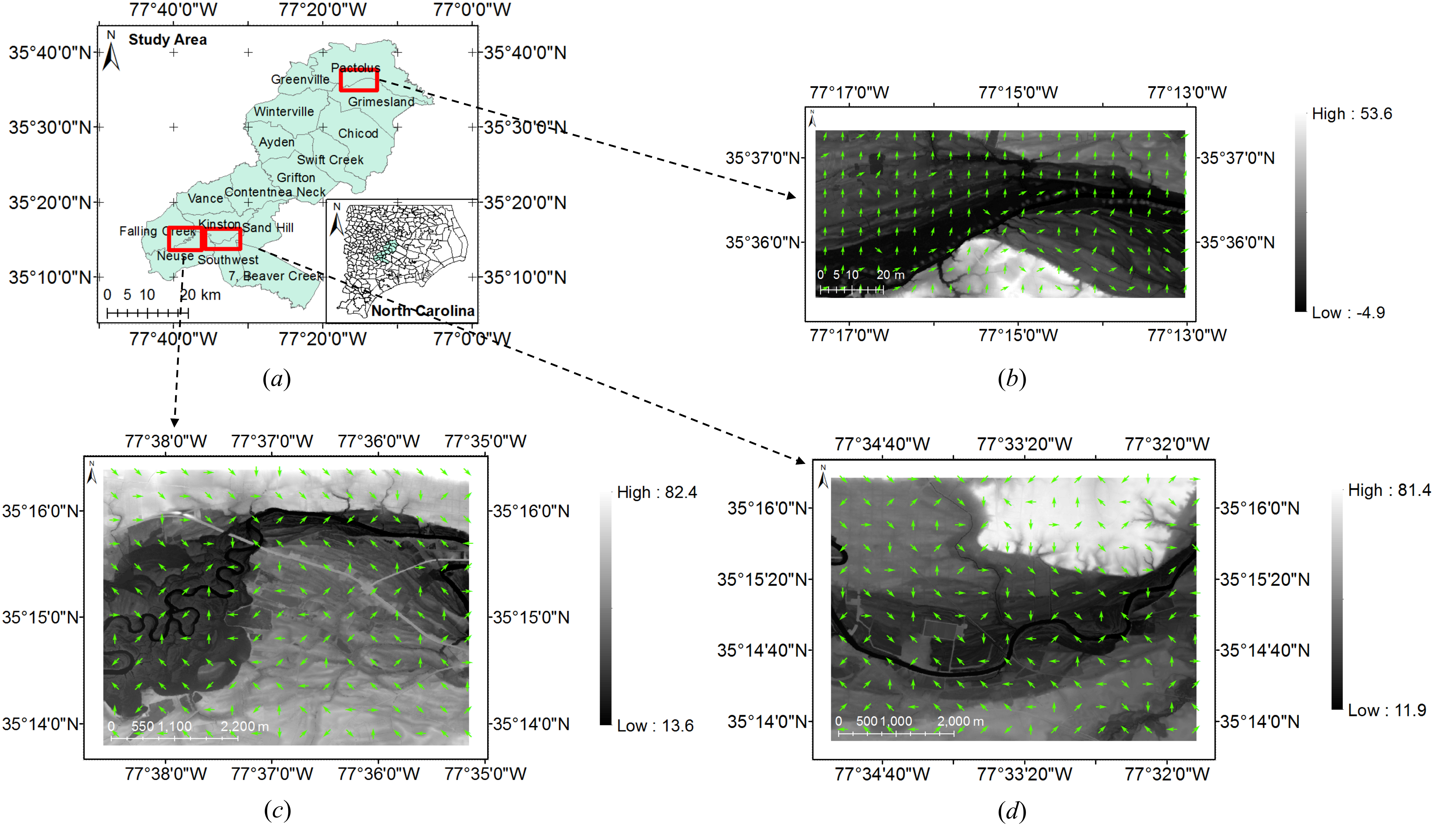}
     \caption{Flow direction maps based on DEM of the three test regions. (\textit{a}) Study Area. (\textit{b}) Grimesland (m). (\textit{c}) Kinston: Area 2 (m). (\textit{d})  Kiston: Area 1 (m) }
    \label{fig:flowdirection}
\end{figure}

We now introduce the HMT model first proposed by~\cite{Xie2018}, which explicitly incorporates the physical constraints of water flow directions based on topography into image classification process. Figure~\ref{fig:hmtrt} provides an illustrative example with eight consecutive locations (pixels) in 1D space. We can consider these eight locations as eight consecutive pixels on a 1D intersection of a DEM layer. Figure~\ref{fig:hmtrt}(a) shows the DEM values of the eight pixel locations with a topography structure. Due to gravity, water tends to flow from a location to nearby lower locations. If location 5 is flooded, then locations 2 to 4 and 6 to 7 must also be flooded. Such dependency structure of pixel classes based on topography can be represented in a tree structure shown by Figure~\ref{fig:hmtrt}(b), whereby any node being flooded indicates all sub-tree nodes being flooded as well. For  example, in Figure~\ref{fig:hmtrt}(b), if  node  5  is  flood,  all  the  subtree  nodes  (2  to  4, and  6  to  7)  must  be  flooded.  Similarly, if node 1 is flooded, nodes 2 to 7 must also be flooded. Though  the  example  shows  the  flow  dependency  tree based on elevation values in one dimensional space (DEM along a 1D line segment), the idea can be easily generalized to a two dimensional DEM map, which is used in this paper. Figure~\ref{fig:flowdirection} shows the topography (illustrated by flow directions) on DEM in the three test regions. It is worth noting that the flow dependency tree structure is not merely based on flow directions between adjacent pixels. For example, if we were only using flow directions between adjacent pixels in Figure~\ref{fig:hmtrt}(a) to construct the flow dependency tree, we can only capture the dependency between pixel $5$ and its neighbors $4$ and $6$ and miss the dependency between pixel $1$ and pixel $5$ (though in reality, if pixel 1 is flood, pixel 5 should also be flood since water can fill into pixel 5 from pixel 1). The flow dependency tree can be efficiently constructed from a DEM layer based on computational topology algorithms (details are in~\cite{Xie2018}). Note that in this paper, we use the word \emph{node}, \emph{location}, \emph{sample}, and \emph{pixel} exchangeably.
%(a) Eight consecutive sample locations in 1D space
%(b) Partial order constraint in a reverse tree
%(c) Hidden Markov tree

% \begin{figure}
%     \centering
%     \subfigure[Study Area]{%
%       \includegraphics[width=2.7in]{figures/FlowDirectionStudyArea.PNG}
% }
% \subfigure[Flow directions on DEM of the test region in Grimesland]{%
%       \includegraphics[width=2.7in]{figures/TC3/TC3FlowDirection.PNG}
% }
% \subfigure[Flow directions on DEM of the first test region in Kinston]{%
%       \includegraphics[width=2.7in]{figures/TC1/TC1FlowDirection.PNG}
% }
% \subfigure[Flow directions on DEM of the second test region in Kinston]{%
%       \includegraphics[width=2.7in]{figures/TC2/TC2FLowDirection.PNG}
% }
%     \caption{Flow direction maps of the three test regions}
%     \label{fig:flowdirection}
% \end{figure}

The HMT model incorporates the class dependency structure from topography into a hidden class layer. As illustrated in Figure~\ref{fig:hmtrt}(c), an HMT model consists of two layers: a hidden class layer of unknown pixel classes (e.g., flood, dry), and an observation layer of sample spectral features (e.g., spectral band values). Each node corresponds to a spatial data sample (image pixel location). Arrow directions between class nodes (with gray shading) capture class transitional probability based on the topography structure. Arrows between an observation node (without shading) and a hidden class node (with gray shading) captures the conditional probability of a pixel's spectral signatures given its class (local likelihood). In this way, the class of a pixel is determined by not only the spectral signatures of itself but also the classes of other pixels based on water flow directions in topography surface. Such model structure is important since the spectral signatures of individual pixels or their local neighborhoods can be misleading due to noise, shadows, obstacles (e.g., tree canopies, clouds), as well as spectral confusion (i.e., the same pixel color may correspond to different classes due to varying illustrations). The class dependency structure  based on the topography constraint provides an opportunity to overcome these issues.

HMT is a probabilistic graphic model. It generalizes the common hidden Markov model from a total order sequence structure to a partial order tree structure. Based on the dependency structure in the model, the joint distribution of all samples' features and classes are expressed as Equation~\eqref{eq:joint}, where $\mathbf{X}$ and $\mathbf{Y}$ are matrices of the spectral features and class labels of all samples respectively, $\mathit{n}$ is the sample (node) index, $\mathit{N}$ is the total number of samples (i.e., nodes, or pixels), $\boldsymbol{x_n}$ and $y_n$ are the spectral feature and class label of the $n$th sample, ${\mathrm{P}_n}$ is the set of parent nodes of the $n$th node in the dependency tree, and $y_{k\in\mathrm{P}_n}\equiv\{y_k|k\in\mathrm{P}_n\}$ is the set of class nodes corresponding to parents of the $n$th node. For a leaf node $\mathit{n}$, ${\mathrm{P}_n} = \emptyset$, and $P(y_n|y_{k\in\mathrm{P}_n})\equiv P(y_n)$.
\begin{dmath}\label{eq:joint}
P(\mathbf{X},\mathbf{Y}) = P(\mathbf{X}|\mathbf{Y})P(\mathbf{Y}) = \prod_{n = 1}^N P(\boldsymbol{x_n}|y_n) \prod_{n = 1}^NP(y_n|y_{k\in\mathrm{P}_n})
\end{dmath}

%\begin{equation}\label{eq:joint2}
%    P(\mathbf{X},\mathbf{Y})= P(\mathbf{X}|\mathbf{Y})P(\mathbf{Y}) = \prod_{n=1}^N %P(\boldsymbol{x}_n|y_n) \prod_{n=1}^NP(y_n|y_{k\in\mathcal{P}_n})
%\end{equation}
Specifically, the conditional probability of a sample feature vector given its class can be assumed i.i.d. Gaussian for simplicity, as shown in Equation~\eqref{eq:featureclassprob}, where $\boldsymbol{\mu}_{y_n}$ and $\boldsymbol{\Sigma}_{y_n}$ are the mean and covariance matrix of feature vector $\boldsymbol{x_n}$ for class $y_n$ ($y_n = 0,1$ with $0$ for dry, and $1$ for flood). It is worth noting that $P(\boldsymbol{x_n}|y_n)$ could be more general than i.i.d. Gaussian. This conditional probability captures the local likelihood of a pixel's class (flood or dry) based on its own spectral features. For example, pixels corresponding to exposed flood water surface often look brown in color. However, for pixels corresponding to noise and obstacles (e.g., tree canopies), their spectral colors may not indicate the true underlying classes. That is why we need to use the class transitional probability to capture the dependency between pixel classes based on topography.
\begin{equation}\label{eq:featureclassprob}
    P(\boldsymbol{x_n}|y_n)\sim \mathrm{N}(\boldsymbol{\mu}_{y_n},\boldsymbol{\Sigma}_{y_n})
\end{equation}

Class transitional probability follows the partial order constraint from topography. For example, due to gravity, if any parent node's class is \emph{dry}, the child's class must be \emph{dry}; if all parents' classes are \emph{flood}, then the child has a high probability of being \emph{flood}. For example, in Figure~\ref{fig:hmtrt}, if node 5 is flood, then nodes 2 to 4 and 6 to 7 must also be flood. Consider \emph{flood} as the positive class (class $1$) and \emph{dry} as the negative class (class value $0$), the transitional probability is actually conditioned on the product of parent classes $y_{\mathrm{P}_n}\equiv\prod_{k\in\mathrm{P}_n}y_k$. The specific transitional probability is expressed in Table~\ref{tab:transitionprob}, where $\rho$ is the probability of a child node being class $1$ (flood) given all parents in class $1$ (flood). Due to spatial autocorrelation, the value of $\rho$ is very high (close to 1). For a leaf node $n$, ${\mathrm{P}_n} = \emptyset$. The transitional probability is degraded into simple class prior probability $P(y_n|y_{k\in\mathrm{P}_n})\equiv P(y_n)$ as expressed on the right of Table~\ref{tab:priorprob}, where $\pi$ is the prior probability of $y_n$ being in class $1$. The class transitional probability is a key component is the HMT model. The local likelihood of individual pixels based on their own spectral features can be misleading due to noise, obstacles, and spectral confusion. The class transitional probability provides additional dependency constraints between pixels' classes based on topography. In this case, even if some pixels' local class likelihood is wrongly estimated, it can still be corrected by other good pixels (whose local class likelihood is more trustworthy).

\begin{table}
\caption{Class transition probability $P(y_n|y_{\mathrm{P}_n})$ }
\label{tab:transitionprob}
\centering
\begin{tabular}{ccc}\hline
 & $y_{\mathrm{P}_n} = 0$ & $y_{\mathrm{P}_n} = 1$\\ \hline
$y_n = 0$ & $1$ & $1-\rho$\\ \hline
$y_n = 1$ & $0$ & $\rho$ \\ \hline
\end{tabular}\hspace{2mm}
\end{table}

\begin{table}
\caption{Class prior probability}
\label{tab:priorprob}
\centering
\begin{tabular}{ccc}\hline
 & $P(y_n)$\\ \hline
$y_n = 0$ & $1-\pi$\\ \hline
$y_n = 1$ & $\pi$ \\ \hline
\end{tabular}
\end{table}

The parameters of HMT include the mean and covariance matrix of sample features in each class, prior probability of leaf node classes, and class transition probability for non-leaf nodes. We denote the entire set of parameters as $\boldsymbol{\Theta} = \{\rho, \pi, \boldsymbol{\mu}_c, \boldsymbol{\Sigma}_c|c = 0,1 \}$. Learning the set of parameters poses two major challenges: first, there exist unknown hidden class variables $\mathbf{Y} = [y_1,...,y_N]^\top$, which are non-i.i.d.; second, the number of samples (nodes) is huge (up to hundreds of millions of pixels). 

To address these challenges, the EM algorithm and message (belief) propagation can be used. The EM-based approach has the following major steps:
\begin{enumerate}
    \item Initialize parameter set $\boldsymbol{\Theta}_0$
    \item Compute posterior distribution of hidden classes:  $P(\mathbf{Y}|\mathbf{X},\mathbf{\Theta_0})$ 
    \item Compute posterior expectation of log likelihood:\\ $\mathrm{LL}(\boldsymbol{\Theta}) = \mathbb{E}_{\mathbf{Y}|\mathbf{X},\boldsymbol{\Theta_0}}\log P(\mathbf{X},\mathbf{Y}|\boldsymbol{\Theta})$
    \item Update parameters:
    $\boldsymbol{\Theta_0}\leftarrow\argmax_{\boldsymbol{\Theta}}\mathrm{LL}(\boldsymbol{\Theta})$\\
    Return $\boldsymbol{\Theta_0}$ if it's converged (no updates), otherwise go back to (2)
\end{enumerate}

Among the four steps above, step (2) that computes the joint posterior distribution of all sample classes $P(\mathbf{Y}|\mathbf{X},\mathbf{\Theta_0})$ is practically infeasible due to the large number of hidden class nodes that are non-i.i.d. Fortunately, it is not necessary to compute the entire joint posterior distribution of all sample classes. In fact, we only need the marginal posterior distribution of a node's and its parents' classes for non-leaf nodes, as well as the marginal posterior distribution of a node's class for leaf nodes. The reason can be explained through the expression of the posterior expectation of log likelihood in Equation~\eqref{eq:postexpll}. Note that for leaf node, $\mathrm{P}_n = \emptyset$, and the last term in the last line of above equation is degraded, i.e., $\log{P(y_n|y_{k\in\mathrm{P}_n},\mathbf{\Theta})}P(y_n,y_{k\in\mathrm{P}_n}|\mathbf{X},\boldsymbol{\Theta_0}) = \log{P(y_n|\boldsymbol{\Theta})}P(y_n|\mathbf{X},\boldsymbol{\Theta_0})$. It can be seen that the joint likelihood can be decomposed into local factors only related to an individual node $n$ and its parents $\mathrm{P}_n$. To compute the expected log likelihood expression, we first need to compute the local marginal posterior distribution such as $P(y_n|\mathbf{X},\boldsymbol{\Theta_0})$ and $P(y_n,y_{k\in\mathrm{P}_n}|\mathbf{X},\boldsymbol{\Theta_0})$.
\begin{equation}\label{eq:postexpll}
\begin{split}
\mathrm{LL}(\boldsymbol{\Theta}) & = \mathbb{E}_{\mathbf{Y}|\mathbf{X},\mathbf{\Theta_0}}\log P(\mathbf{X},\mathbf{Y}|\boldsymbol{\Theta})\\
& = \mathbb{E}_{\mathbf{Y}|\mathbf{X},\boldsymbol{\Theta_0}}\log\left\{ \prod_{n = 1}^N P(\boldsymbol{x}_n|y_n,\boldsymbol{\Theta}) \prod_{n = 1}^NP(y_n|y_{k\in\mathrm{P}_n},\boldsymbol{\Theta})\right\}\\
% &=\int_\mathbf{Y}d\mathbf{Y}\\
& = \sum\limits_{\mathbf{Y}}{ P(\mathbf{Y}|\mathbf{X},\boldsymbol{\Theta_0})
\sum_{n = 1}^{N}{\left\{\log{P(\boldsymbol{x}_n|y_n,\boldsymbol{\Theta})}+\log{P(y_n|y_{k\in\mathrm{P}_n},\boldsymbol{\Theta})}\right\}}}\\
& = \sum_{n = 1}^{N}\sum_{y_n}P(y_n|\mathbf{X},\boldsymbol{\Theta_0})\log{P(\boldsymbol{x}_n|y_n,\boldsymbol{\Theta})}\\
&\quad\quad+\sum_{n = 1}^{N}~\sum_{y_n,y_{k\in\mathrm{P}_n}}P(y_n,y_{k\in\mathrm{P}_n}|\mathbf{X},\boldsymbol{\Theta_0}) \log{P(y_n|y_{k\in\mathrm{P}_n},\boldsymbol{\Theta})}\\
\end{split}
\end{equation}
%Though we introduce our HMT in the context of flood mapping, the model can potentially be used for a broad class of classification problems in which class labels follow a partial order dependency. Examples include predicting pollutants in river stream networks and traffic congestion in road networks.

To compute the marginal posterior distribution $P(y_n,y_{k\in\mathrm{P}_n})$ and $P(y_n)$ (we omit the condition on $\mathbf{X}$ and $\boldsymbol{\Theta_0}$ for brevity), we can use the message propagation method based on the sum and product algorithm. Due to space limit, we do not provide the detailed math formula. The intuition of message propagation along graph (or tree) nodes is a process of marginalizing out those corresponding node variables in a joint distribution. The marginalization process is done over a tree by a traversal order to reduce redundant computation. More details can be found in~\cite{kschischang2001factor,ronen1995parameter}.

After learning model parameters, we can infer hidden class variables by maximizing the overall probability. A naive approach that enumerate all combinations of class assignment is infeasible due to the exponential cost. We use a dynamic programming based method called \emph{max-sum} from~\cite{rabiner1989tutorial}. The process is similar to the sum and product algorithm above. The main difference is that instead of using sum operation, we need to use max operation in message propagation, and also memorize the optimal variable values. We omit the details due to space limit. The main intuition is that in class inference, the class label for each node is not only based on the likelihood of its own spectral features, but also other nodes' class labels through transitional probability. In other words, the class labels of all nodes are jointly inferred based on the topography constraint. 

Note that our HMT model uses transdutive learning, i.e., the model is trained based on not only the features and labels from the training pixels but also the features from the test samples. This is reflected by the fact that the model will create a flow dependency tree basd on the topography of the test region. Thus, the model needs to be re-trained for each specific test region. This is different from many existing machine learning algorithms based on inductive learning: once a general model is learned from training pixels, it can be applied to any test set instead of only being applied to a particular test set. Since the computational structure of both the parameter learning and class inference is largely based on tree traversal operations, the time cost is linear to the number of tree nodes. The model training is computationally very fast. 
\section{Results}\label{sec:results}
%{\color{red} Arpan: please put results here}
In this section, we compared the HMT model with several baseline methods in classification performance on three real-world datasets. Unless specified otherwise, we used default parameters in open source tools for baseline methods. Candidate classification methods include: 
\begin{itemize}
    \item {\bf Non-spatial classifiers}: We tested random forest ({\bf RF}) and gradient boosted tree ({\bf GBM}) in R packages on spectral features (red, green, blue spectral bands). 
    \item {\bf Non-spatial classifier with post-processing label propagation (LP):} We tested {\bf RF} and {\bf GBM} together with label propagation smoothing using 4 neighborhood~\cite{zhu2002learning}. Label propagation was added on top of the pre-classified map from a classifier through updates based on neighborhood majority.
    \item {\bf HMT}: We used the model from~\cite{Xie2018} implemented in C++. We added an additional hidden class layer to accommodate the large scale of tree canopies overlaying flood water.
\end{itemize}

The evaluation metrics we used include precision ($P$) (user's accuracy), recall ($R$) (producer's accuracy), and F-score ($F_1$) (the harmonic mean between precision and recall). The definition of the metrics are provided in Equation~\ref{eq:prec}, Equation~\ref{eq:recall} and Equation~\ref{eq:fscore}, where TP, FP, and FN are the numbers of true positives (TP) (i.e., pixels truly in the positive class and correctly predicted into the positive class), false positives (FP) (i.e., pixels truly in the negative class but mistakenly predicted into the positive class) and false negatives (FN) (i.e., pixels truly in the positive class but mistakenly predicted into the negative class) respectively. A high precision on a class means that among the pixels predicted into the class, most predictions are correct. A high recall on a class means that among the pixels that are actually in the class category, most of them are correctly identified in the prediction. Quantitative evaluation results on classification accuracy was based on the predictions on randomly sampled pixels from labeled test polygons.
\begin{equation}\label{eq:prec}
    \text{$P$} = \frac{\text{TP}}{\text{TP+FP}}
\end{equation}
\begin{equation}\label{eq:recall}
    \text{$R$} = \frac{\text{TP}}{\text{TP+FN}}
\end{equation}
\begin{equation}\label{eq:fscore}
   % \text{$F_1$} = \frac{2 \times \text{$P$}*\text{$R$}}{\text{$P$}+\text{$R$}}
   \text{$F_1$} = 2 \times \frac{P \times R}{P + R} 
\end{equation}

% \emph{Dataset description:} We used three flood mapping datasets from the cities of Grimesland and Kinston in North Carolina during Hurricane Mathew in 2016. Explanatory features were red, green, blue bands in aerial imagery from NOAA National Geodetic Survey~\cite{ngs}. The potential field was digital elevation map from the University of North Carolina Libraries~\cite{ncsudem}. All data were resampled into 2 meter by 2 meter resolution. The number of training and test samples were $5000$ per class (flood, dry) in both datasets. The test region size was  1856 by 4104 in Grimesland and 2313 by 3207 in Area 1 in Kinston and 2418 by 3205 in Area 2 in Kinston. Training samples in Grimesland were drawn beyond but close to the test region. Training samples in Grimesland dataset were drawn far away from the test region to evaluate model generalizability. 

\subsection{Classification Performance Comparison}
% Results on the Grimesland dataset were summarized in Table~\ref{tab:classificationComp3}. Decision tree, random forest, gradient boosted tree, and maximum likelihood classifier achieved overall F-scores between 0.60 and 0.75 on raw features. Post-processing based on label propagation (LP) slightly improved performance of non-spatial classifiers due to the removal of salt-and-pepper noise errors. However, the performance was still quite inferior compared to our hidden Markov tree, which achieved around 0.99 F-score on both classes. 
The comparison of classification performance between different methods on the three datasets are summarized in Table~\ref{tab:classificationComp1}, Table~\ref{tab:classificationComp2}, and Table~\ref{tab:classificationComp3} respectively.

%TC 3,  
\begin{table}
\centering
\caption{Classification on real dataset in Grimesland NC} %Area2 
\label{tab:classificationComp1}
\begin{tabular}{cccccc}
\hline
Classifier & Class & $P$ & $R$ & $F_1$ & Average $F_1$\\ \hline
%section1: non-spatial ensemble
%\multirow{2}{*}{DT}&Dry&{0.52}&{0.85}&{0.65}&\multirow{2}{*}{0.69}\\ 
% &Flood&{0.89}&{0.61}&{0.72}&\\ \hline
 
\multirow{2}{*}{RF}&Dry&{0.45}&{0.96}&{0.61}&\multirow{2}{*}{0.60}\\ 
 &Flood&{0.95}&{0.42}&{0.58}&\\ \hline
 
 \multirow{2}{*}{GBM}&Dry&{0.51}&{0.86}&{0.64}&\multirow{2}{*}{0.68}\\ 
 &Flood&{0.90}&{0.59}&{0.71}&\\ \hline 
 
%\multirow{2}{*}{MLC}&Dry&{0.59}&{0.85}&{0.70}&\multirow{2}{*}{0.75}\\ 
% &Flood&{0.90}&{0.71}&{0.80}&\\ \hline 

%\multirow{2}{*}{DT+LP}&Dry&{0.57}&{0.96}&{0.71}&\multirow{2}{*}{0.73}\\ 
% &Flood&{0.97}&{0.62}&{0.75}&\\ \hline 
 
\multirow{2}{*}{RF+LP}&Dry&{0.45}&{0.99}&{0.62}&\multirow{2}{*}{0.59}\\ 
 &Flood&{0.99}&{0.40}&{0.57}&\\ \hline 
 
 \multirow{2}{*}{GBM+LP}&Dry&{0.53}&{0.97}&{0.68}&\multirow{2}{*}{0.70}\\ 
 &Flood&{0.97}&{0.57}&{0.72}&\\ \hline  
 
 %\multirow{2}{*}{MLC+LP}&Dry&{0.64}&{0.92}&{0.75}&\multirow{2}{*}{0.79}\\ 
 %&Flood&{0.95}&{0.74}&{0.83}&\\ \hline  
 
\multirow{2}{*}{HMT}&Dry&{0.99}&{0.98}&{0.99}&\multirow{2}{*}{0.99}\\ 
 &Flood&{0.99}&{0.99}&{0.99}&\\ \hline 

\end{tabular}
\end{table}
%TC 1,  first dataset
\begin{table}
\centering
\caption{Classification on real dataset in Kinston NC: Area 1} %Area1
\label{tab:classificationComp2}
\begin{tabular}{cccccc}
\hline
Classifier & Class & $P$ & $R$ & $F_1$ & Average $F_1$\\ \hline
%section1: non-spatial ensemble
%\multirow{2}{*}{DT}&Dry&{0.87}&{0.89}&{0.88}&\multirow{2}{*}{0.80}\\ 
% &Flood&{0.74}&{0.70}&{0.72}&\\ \hline
 
\multirow{2}{*}{RF}&Dry&{0.84}&{0.96}&{0.90}&\multirow{2}{*}{0.79}\\ 
 &Flood&{0.87}&{0.57}&{0.69}&\\ \hline
 
 \multirow{2}{*}{GBM}&Dry&{0.86}&{0.88}&{0.87}&\multirow{2}{*}{0.79}\\ 
 &Flood&{0.72}&{0.67}&{0.70}&\\ \hline 
 
%\multirow{2}{*}{MLC}&Dry&{0.93}&{0.78}&{0.84}&\multirow{2}{*}{0.79}\\ 
% &Flood&{0.64}&{0.86}&{0.73}&\\ \hline 

%\multirow{2}{*}{DT+LP}&Dry&{0.88}&{0.96}&{0.92}&\multirow{2}{*}{0.85}\\ 
% &Flood&{0.87}&{0.70}&{0.78}&\\ \hline 
 
\multirow{2}{*}{RF+LP}&Dry&{0.84}&{0.99}&{0.91}&\multirow{2}{*}{0.82}\\ 
 &Flood&{0.97}&{0.58}&{0.72}&\\ \hline 
 
 \multirow{2}{*}{GBM+LP}&Dry&{0.87}&{0.96}&{0.91}&\multirow{2}{*}{0.83}\\ 
 &Flood&{0.87}&{0.66}&{0.75}&\\ \hline  
 
 %\multirow{2}{*}{MLC+LP}&Dry&{0.94}&{0.82}&{0.88}&\multirow{2}{*}{0.83}\\ 
 %&Flood&{0.70}&{0.88}&{0.77}&\\ \hline  
 
\multirow{2}{*}{HMT}&Dry&{1.00}&{0.94}&{0.96}&\multirow{2}{*}{0.95}\\ 
 &Flood&{0.87}&{0.99}&{0.93}&\\ \hline 
\end{tabular}
\end{table}
%TC 2,  
\begin{table}
\centering
\caption{Classification on real dataset in Kinston NC: Area 2} %Area2 
\label{tab:classificationComp3}
\begin{tabular}{cccccc}
\hline
Classifier & Class & $P$ & $R$ & $F_1$ & Average $F_1$\\ \hline
%section1: non-spatial ensemble
%\multirow{2}{*}{DT}&Dry&{0.62}&{0.88}&{0.73}&\multirow{2}{*}{0.68}\\ 
% &Flood&{0.82}&{0.52}&{0.64}&\\ \hline
 
\multirow{2}{*}{RF}&Dry&{0.57}&{0.96}&{0.72}&\multirow{2}{*}{0.61}\\ 
 &Flood&{0.90}&{0.36}&{0.51}&\\ \hline
 
 \multirow{2}{*}{GBM}&Dry&{0.61}&{0.84}&{0.72}&\multirow{2}{*}{0.67}\\ 
 &Flood&{0.83}&{0.50}&{0.62}&\\ \hline 
 
%\multirow{2}{*}{MLC}&Dry&{0.70}&{0.91}&{0.79}&\multirow{2}{*}{0.77}\\ 
% &Flood&{0.89}&{0.64}&{0.75}&\\ \hline 

%\multirow{2}{*}{DT+LP}&Dry&{0.63}&{0.94}&{0.76}&\multirow{2}{*}{0.71}\\ 
% &Flood&{0.91}&{0.51}&{0.65}&\\ \hline 
 
\multirow{2}{*}{RF+LP}&Dry&{0.56}&{0.98}&{0.71}&\multirow{2}{*}{0.59}\\
 &Flood&{0.94}&{0.31}&{0.46}&\\ \hline 
 \multirow{2}{*}{GBM+LP}&Dry&{0.62}&{0.95}&{0.75}&\multirow{2}{*}{0.69}\\ 
 &Flood&{0.91}&{0.48}&{0.63}&\\ \hline  
% \multirow{2}{*}{MLC+LP}&Dry&{0.70}&{0.96}&{0.81}&\multirow{2}{*}{0.78}\\ 
% &Flood&{0.94}&{0.64}&{0.76}&\\ \hline  
\multirow{2}{*}{HMT}&Dry&{1.00}&{0.92}&{0.96}&\multirow{2}{*}{0.96}\\ 
 &Flood&{0.93}&{1.00}&{0.97}&\\ \hline 
\end{tabular}
\end{table}

From the results on Grimesland dataset in Table~\ref{tab:classificationComp1}, we can see that traditional classification models such as random forest and gradient boosted trees performance poorly (with overall $F_1$ scores of 0.6 and 0.68). The gradient boosted model is slightly better than random forest. Within each method, the precision for the flood class is very high (0.95 and 0.9), meaning that the flood pixels being identified are largely correct. These correctly predicted flood pixels are mostly open exposed flood surface. However, the recall of the flood class is very low (0.42 and 0.59), meaning that there are a significant amount (around half) of flood pixels not being identified. The large amount of actual flood pixels being missed in the prediction results are largely due to the obstacles in the image (e.g., tree canopies). The spectral signature of the tree canopies overlaying flood water is very close to that of the trees in the dry area, and thus they are mistakenly classified into the dry class (though there are flood water beneath).  Adding post-processing through label propagation slightly decreases the $F_1$ of random forest (from 0.6 to 0.59) but slightly increases the $F_1$ of gradient boosted trees (from 0.68 to 0.7). The reason is that neighborhood smoothing could reduce salt-and-pepper noise errors, but also could mistakenly change a correctly predicted pixel if most of the neighbors were mis-classified.
In contrast, the HMT model performed the best, with an $F_1$ around 0.99. It almost perfectly identified the flood extent. The results of the Area 2 in the city of Kinston are summarized in Table~\ref{tab:classificationComp3}. The trends look very close to the results in Table~\ref{tab:classificationComp1}.

The results on the dataset collected from the Area 1 in the city of Kinston are summarized in Table~\ref{tab:classificationComp2}. The overall comparison between methods looks similar to the results on the other two datasets. Some differences exist though. The classification performance of non-spatial classifiers is generally better compared with the results on the other datasets, with the $F_1$ scores of random forest and gradient boosted model around 0.8. The precision on the flood class is generally worse than other datasets, while the recalls is generally better. This means that the non-spatial models identified more flood pixels on this dataset compared with the other datasets, but the accuracy is lower among those flood pixels being identified. We also observe that HMT performed slightly worse on this dataset than other dataset, largely due to a lower precision on the flood class (0.87). HMT is still significantly more accurate than the other methods on this dataset.

\subsection{Interpretation of Results}
\begin{figure}
    \centering
    \includegraphics[width=6.0in]{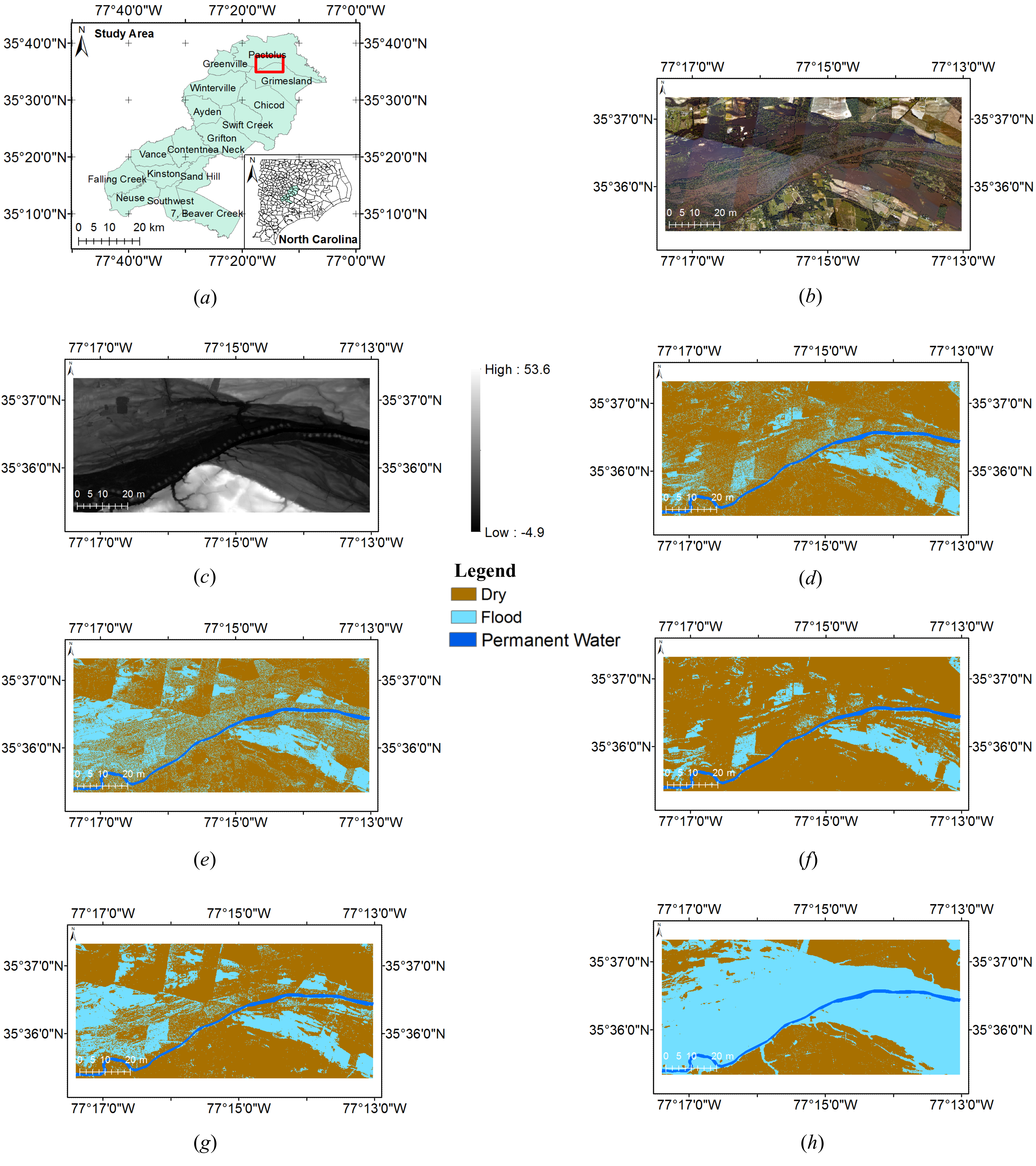}
    \caption{Results on Matthew flood, Grimesland, NC. (\textit{a}) Study area. (\textit{b}) High-resolution aerial imagery in Grimesland (Source: NOAA National Geodesic Survey, Date: 15 October 2016). (\textit{c}) DEM (m) (\textit{d}) RF result (\textit{e}) GBM result (\textit{f}) RF+LP result (\textit{g}) GBM+LP result (\textit{h}) HMT result}
    \label{fig:map1}
\end{figure}

\begin{figure}
    \centering
    \includegraphics[width=6.0in]{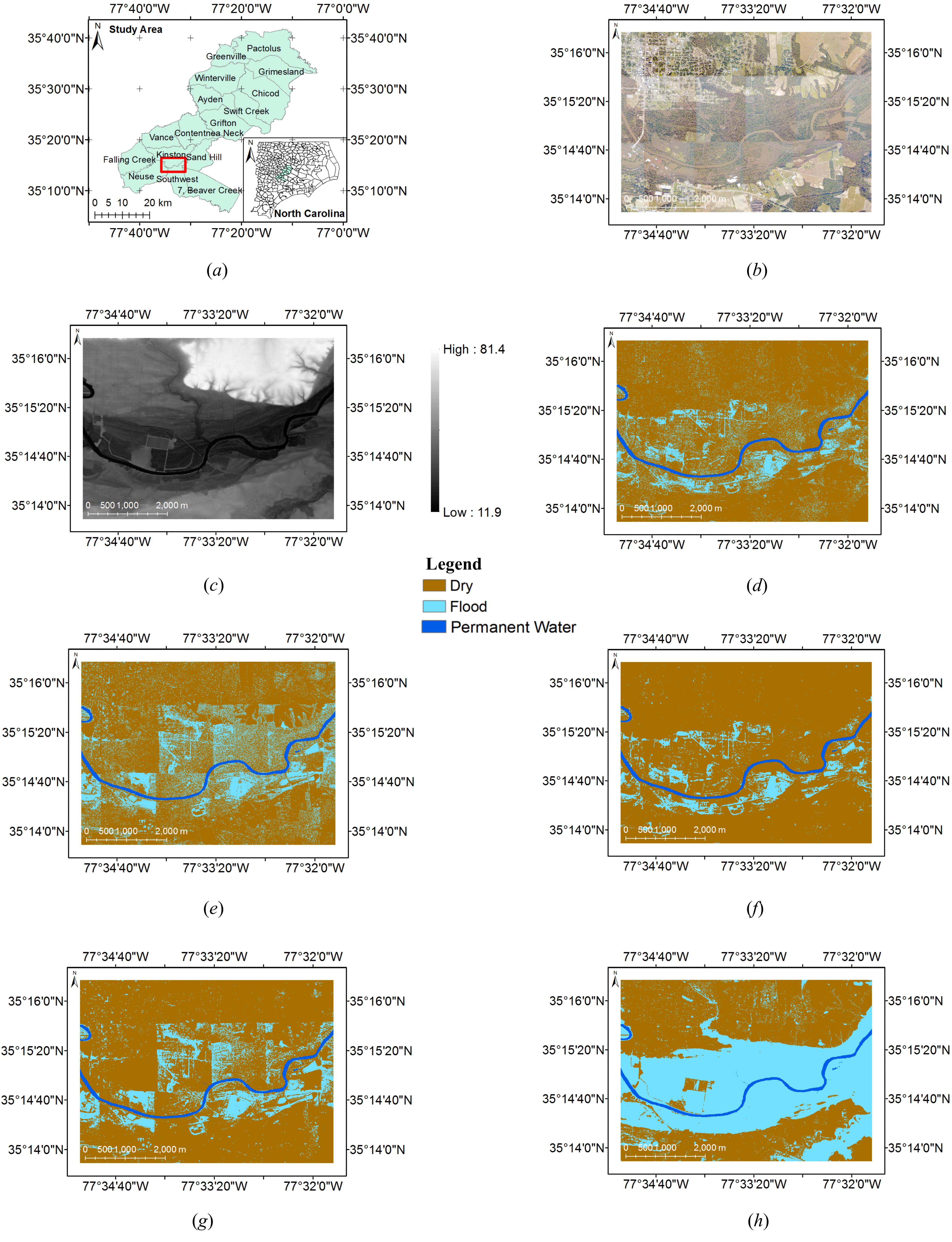}
    \caption{Results on Matthew flood, Area 1 of Kinston, NC.  (\textit{a}) Study area. (\textit{b}) High-resolution aerial imagery in Kinston: Area 1 (Source: NOAA National Geodesic Survey, Date: 15 October 2016). (\textit{c}) DEM (m) (\textit{d}) RF result (\textit{e}) GBM result (\textit{f}) RF+LP result (\textit{g}) GBM+LP result (\textit{h}) HMT result}
    \label{fig:map2}
\end{figure}

\begin{figure}
    \centering
    \includegraphics[width=6.0in]{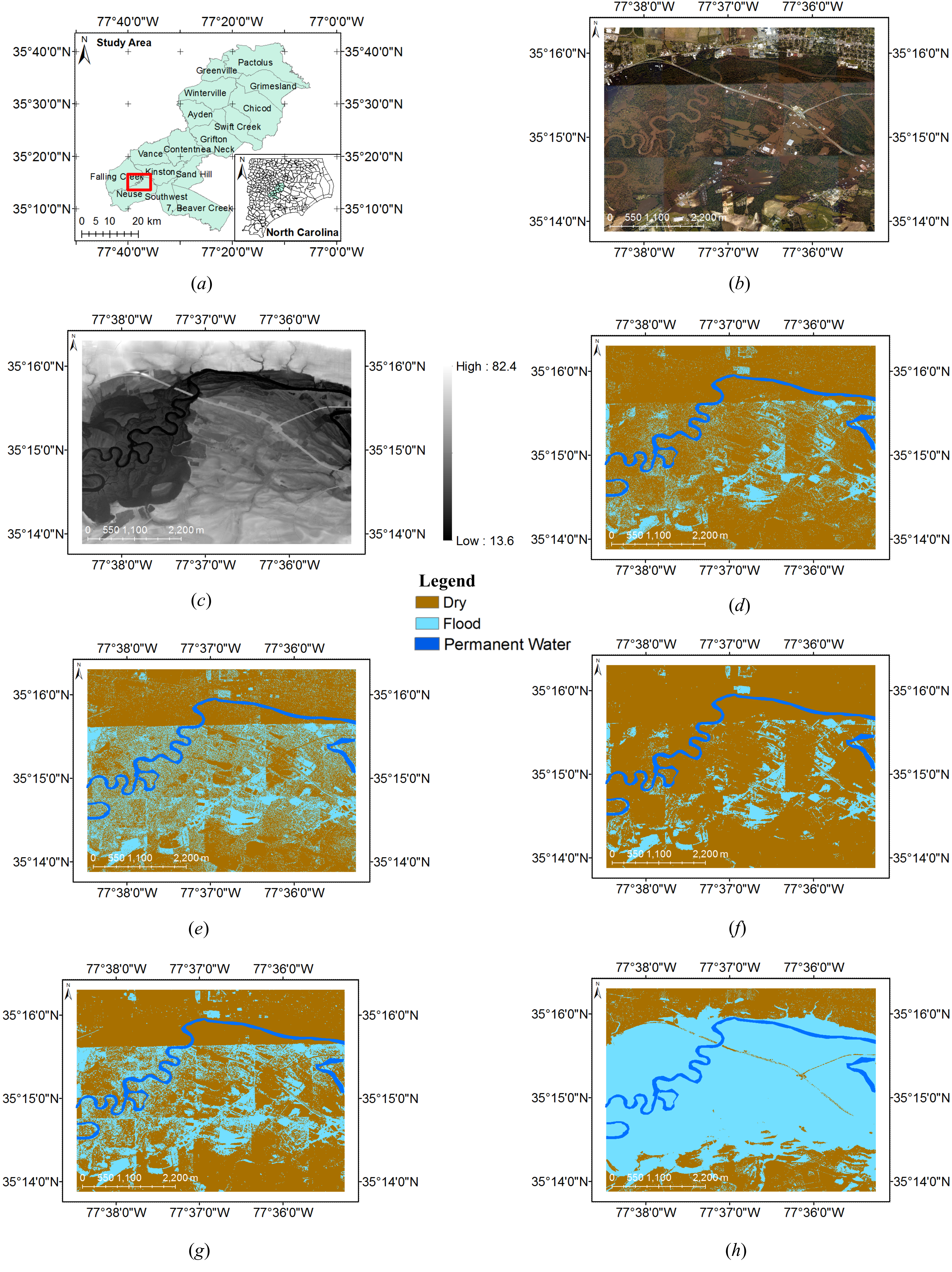}
    \caption{Results on Matthew flood, Area 2 of Kinston, NC. (\textit{a}) Study area. (\textit{b}) High-resolution aerial imagery in Kinston: Area 2 (Source: NOAA National Geodesic Survey, Date: 15 October 2016). (\textit{c}) DEM (m) (\textit{d}) RF result (\textit{e}) GBM result (\textit{f}) RF+LP result (\textit{g}) GBM+LP result (\textit{h}) HMT result}
    \label{fig:map3}
\end{figure}
We now provide detailed interpretation of classification results through visualized flood maps from different methods. The input aerial imagery and digital elevation model, as well as classified flood maps from methods on three datasets are shown in Figure~\ref{fig:map1}, Figure~\ref{fig:map2}, and Figure~\ref{fig:map3} respectively. 

Figure~\ref{fig:map1} shows the classified flood maps on the Grimesland dataset (flood is shown in blue while dry is shown in brown). The study area is shown in Figure~\ref{fig:map1} (a) and the input aerial imagery with red, green, and blue bands are shown in Figure~\ref{fig:map1} (b). The elevation map is shown in Figure~\ref{fig:map1} (c), with black color indicating low elevation and white color indicating high elevation. From the aerial imagery, we can found that the central part of the region is largely flooded (black and brown colors). We can also observe that there are a significant amount of vegetation (tree canopies in green color) among the flooded area. These tree canopies have the same color and texture (spectral signature) as the tree canopies in the dry area. Moreover, the aerial imagery has various illumination condition (the same flood surface shows discontinuity in color tones). All these issues pose significant challenges in identifying the correct flood extent. From Figure~\ref{fig:map1} (d)-(e), we can see that RF and GBM successfully identified part of exposed flood surface on the  lower right part of the region. There are some part of exposed flood surface in the lower left part not being identified, probably because the colors of exposed flood pixels in two parts are different due to varying illumination. We also observed that the two models almost all missed the flooded pixels obscured by significant tree canopies. This is because the spectral signatures of the tree canopies overlaying flood water are very close to those signatures in dry areas. This observation is consistent with the low recall observed in Table~\ref{tab:classificationComp1}. In addition, we observe a significant amount of salt-and-pepper noise in the classified maps, due to the fact that both models classify each pixel independently (per-pixel classification) without incorporating the spatial autocorrelation effect. Figure~\ref{fig:map1} (f)-(g) show the results after conducting post-processing through neighborhood label propagation. We can see that post-processing can significantly reduce the amount of salt-and-pepper noise due to the incorporation of neighborhood context. But post-processing with neighborhood level label propagation could not address the errors in missing a large area of flood areas under tree canopies because the errors are systematic beyond a neighborhood scale. Finally, we can see from Figure~\ref{fig:map1} (h) that HMT successfully identified most of the flooded areas, including the areas under tree canopies. The reason is that HMT incorporate the spatial structural dependency between pixel locations based on the topography. For example, the exposed flood surface on the lower right part has relatively higher elevation on the elevation map in Figure~\ref{fig:map1} (c). These pixels have a higher elevation than the other flooded areas obscured by tree canopies. Thus, the model is able to fill the gap of missed flood areas under tree canopies since the topography indicates water flow directions.

The results on Area 1 of Kinston are shown in Figure~\ref{fig:map2}. The study area is shown in Figure~\ref{fig:map2} (a). From the aerial imagery and elevation map in Figure~\ref{fig:map2} (b)-(c), we can observe that the central part of the region is flooded (with brown flood water). There are a significant amount of tree canopies overlaying flood water. The upper and lower parts are mostly dry. There are also varying illuminations within the region (similar to the Grimesland dataset). This is shown in several vertical strips in the aerial images. From the classified maps of random forest and gradient boosted model, we can observe a significant among of salt-and-pepper noise and missed flood areas due to tree canopies as well as spectral confusion due to varying illumination. Adding post-processing reduces the salt-and-pepper noise, as shown in Figure~\ref{fig:map2} (f)-(g). Finally, the HMT model identifies more flood areas with tree canopies. 

The results on Area 2 of Kinston are shown in Figure~\ref{fig:map3}. The study area is shown in Figure~\ref{fig:map3} (a). From the aerial image and the elevation map, we can see similar challenges to the other two datasets. What is more significant in this region is the varying illumination between different strips of the aerial imagery. We can observe a clear horizontal line separating the top part (much darker color) from the central part (lighter color). This is also reflected by the artifacts (horizontal flood boundary) in the classified flood maps from random forest and gradient boosted model (Figure~\ref{fig:map3} (d) to (e)). The results of the two non-spatial models miss a significant amount of flood areas, but gradient boosted model is slightly better in identifying flood. This is consistent with the low recalls of the two methods in Table~\ref{tab:classificationComp3} (with recall slightly higher in gradient boosted model). Post-processing reduces salt-and-pepper noise, but still could not correct the missing flood areas. In contrast, HMT can identify most of the flood areas due to incorporating the topography. 

Note that the HMT prediction output is the entire inundation extent, including the perennial water surface. Perennial water can be separated from the flood water through further post-processing, e.g., change detection. Our reported classification accuracy metrics were based on test polygons that were drawn from the flood area outside the perennial water.

\subsection{Computational Time Costs}
We also conducted computational experiments on the three test regions to measure the time costs of the tree construction, parameter learning and class inference process. Experiments were conducted on a Dell workstation with Intel(R) Xeon(R) CPU E5-2687w v4 @ 3.00GHz, 64GB main memory, and Windows 10. As mentioned in the earlier, our HMT model was implemented in C++. The codes ran sequentially on one CPU core only. Table~\ref{tab:comp} summarizes the results. The number of pixels in three test regions are around 7.6 million, 7.4 million and 7.7 million respectively. The tree construction part only took around 3 seconds. The class inference was also very fast, only taking less than 2 seconds for over 7 million pixels. The parameter learning part contributed to the vast majority of the time costs (around 260 seconds). The reason is that the parameter learning step involves multiple iterations, each of which has message propagation over the entire tree. Overall, the HMT algorithm was highly efficient. It could process over 7 million pixels within 5 minutes. Please note that we did not compare the computational time costs of HMT with those of the baseline models such as GBM and RF, since these baseline models were implemented in a different programming language (R instead of C++) and they did not capture the topography in the classification process.

\begin{table}[h]
    \centering
    \caption{Computational time costs for three study areas}
    \label{tab:comp}    
    \begin{tabular}{ccccc}\hline
    Study area & Size (pixels) & Tree construction (s) & Learning (s)& Inference (s) \\ \hline
    Grimesland & $1856 \times 4104$ &2.88 & 264.57 & 1.94 \\ \hline
    Kinston 1 &  $2313  \times 3207$ &3.00 & 257.75 &1.73  \\ \hline
    Kinston 2 &  $2418  \times 3205$ &3.26 &268.25  &1.93  \\ \hline
    \end{tabular}
\end{table}
\section{Discussions}\label{sec:discuss}
This paper introduces a machine learning approach called hidden Markov tree for flood extent mapping from high-resolution optical imagery and DEM data. Existing image classification algorithms for flood mapping often uses spectral and spatial features at the pixel or neighborhood level, and thus cannot fully address the issues of noise and large scale obstacles (e.g., a large area of vegetation over flood water). In contrast, HMT model captures the topography constraint of water flow directions over the entire study area. Water flow directions are incorporated as a directed tree structure in a hidden class layer. In this way, the order of class labels of pixels (flood or dry) are constrained by the topography. Thus, the model can automatically infer the class labels of pixels not only based on its local (or neighborhood) spectral signatures, but also based on other pixels information in a topography surface. Results show that the model can fill the gap of missing flood extent due to dense vegetation. There exist other ways of incorporating topography into flood extent mapping, e.g., adjusting thresholds on elevation map to find the best matching of flood extent (e.g., \cite{wang2002efficient}). But the HMT model can add value in learning and inferring class labels automatically without the need of manually tuning the thresholds. The HMT model provides a clear statistical framework that makes the model easily interpretable. In addition, the HMT model is also very fast due to the linear time complexity. Classifying the entire area with over 7 million pixels took just a few minutes.

One limitation of the HMT approach is the flow dependency tree derived from the DEM data is quite simple. It assumes that the inundation surface is flat  (with equal elevation). Such assumption is valid for river floodplain with relatively flat topography. When the study area is a big watershed with complex inters of flow direction, the flow dependency tree structure in HMT needs to be re-designed to reflect the complex topography (i.e., to allow for non-flat flood surface). Theoretically, the HMT model can work with any flow direction dependency structure as long as it can be represented in a tree or polytree. Otherwise, we may choose to divide the big watershed into smaller pieces such that the flood surface in each piece can be considered as flat. In future work, we plan to address the limitation by integrating a more sophisticated topography model into HMT for vast watershed areas.

\section{Conclusion and Future Work}
This paper conducts several case studies on using a spatial machine learning model called hidden Markov tree for flood extent mapping based on high-resolution aerial imagery and DEM. The model can automatically incorporate topography constraint (water flow directions) in the image classification process in an efficient manner. Incorporating topography constraint is critically important in addressing noise and obstacles (e.g., tree canopies blocking the view of flood water) in high-resolution imagery. Results show that the HMT model significantly improves several existing methods in classification accuracy (with $F_1$ increasing from around 70 to 80\% to over 95\%) on three study areas in hurricane Matthew floods in North Carolina 2016. 

In future work, we plan to further test the generalizability of the HMT model on datasets collected by different remote sensors (e.g., radar image, MODIS or Landsat imagery) in different flood events and study areas. The HMT model can also be potentially improved by adding morphological characters of a watershed (in addition to spectral bands from optical imagery) as model inputs. We plan to evaluate the HMT approach in refining pre-classified flood maps from other sensors (e.g., MODIS Flood Water maps, which have coarser spatial resolution and contain gaps due to feature obstacles such as tree canopies). We will also explore the integration of the HMT with the recent deep learning models.

\section*{Acknowledgement}
This material is based upon work supported by the NSF under Grant No. IIS-1850546, IIS-2008973, CNS-1951974 and the University Corporation for Atmospheric Research (UCAR).

\section*{Declaration of Interest}
No potential conflict of interest was reported by the authors.

\bibliographystyle{tfcad}
\bibliography{ref}
\listoffigures

\listoftables
\end{document}